# PIQI: Perceptual Image Quality Index based on Ensemble of Gaussian Process Regression

Nisar Ahmed[a], Hafiz Muhammad Shahzad Asif[b], Hassan Khalid[c]

[a,b] Department of Computer Science and Engineering, University of Engineering and Technology, Lahore, Pakistan.
[c] Space and Upper Atmosphere Research Commission, Pakistan.

**Abstract**

Digital images contain a lot of redundancies, therefore, compression techniques are applied to reduce the image size without the loss of reasonable image quality. Same become more prominent in the case of videos which contains image sequences and higher compression ratios are achieved in low throughput networks. Assessment of quality of images in such scenarios has become of particular interest. Subjective evaluation in most of the scenarios is infeasible so objective evaluation is preferred. Among the three objective quality measures, full-reference and reduced-reference methods require an original image in some form to calculate the image quality which is unfeasible in scenarios such as broadcasting, acquisition or enhancement. Therefore, a no-reference Perceptual Image Quality Index (PIQI) is proposed in this paper to assess the quality of digital images which calculates luminance and gradient statistics along with mean subtracted contrast normalized products in multiple scales and color spaces. These extracted features are provided to a stacked ensemble of Gaussian Process Regression (GPR) to perform the perceptual quality evaluation. The performance of the PIQI is checked on six benchmark databases and compared with twelve state-of-the-art methods and competitive results are achieved. The comparison is made based on RMSE, Pearson and Spearman's correlation coefficients between ground truth and predicted quality scores. The scores of 0.0552, 0.9802 and 0.9776 are achieved respectively for these metrics on CSIQ database. Two cross-dataset evaluation experiments are performed to check the generalization of PIQI.

## 1. Introduction

In most of the applications, the quality of an image is momentous and a number of techniques are currently available for the evaluation of image quality [1-5]. There are mainly two classes of such evaluation techniques, the first class requires reference images [6-8] while the second class does not require any reference image [9]. In most of the cases, reference images are not readily available such as in broadcasting, image acquisition and enhancement. Therefore, image quality evaluation without reference images is of paramount importance. The perceptual quality of an image is more important than the change in comparison to original image. The perceptual Image Quality Assessment (IQA) is divided into two categories: subjective evaluation by the humans and objective evaluation through the algorithms [10-12]. Subjective evaluation is the ultimate method of quality assessment as humans are the eventual consumer of digital images. Objective assessment techniques are less expensive, less time consuming and convenient to use and therefore is receiving more focus of the research community [4, 10-13].

There are three categories of IQA models. No-reference (NR) approaches, Full-Reference (FR) approaches and Reduced-Reference (RR) approaches [11]. The FR methods uses the pristine image as a reference and checks it for similarity with the distorted image. The RR techniques are designed for applications such as communication or broadcasting where availability of the reference image is difficult, so a reduced metric of the reference image is transmitted with the original image and comparison is made with this metric rather than the original image. This approach covers some applications with few limitations, but image acquisition and enhancement applications still need NR methods. The no-reference IQA in their early stages assumes that the quality of an image is affected by a certain type of distortion i.e. noise, ringing, blurriness, and compression [10]. Earlier research in the area of no-reference quality assessment, therefore, focus on distortion-aware techniques and use models which estimates the amount of certain distortions in the image [1, 14-16].

Distortion un-aware techniques focus on general-purpose image quality assessment and with no-presumed distortion information. These techniques calculate Natural Scene Statistics (NSS) based features which are inspired by Human Visual System (HVS) and its perception of visual quality [17, 18]. Mean Opinion Score (MOS) of natural images is obtained from subjective evaluation experiments and used for supervised training of regression algorithms [19]. Performance of these methods therefore largely depends on the feature set which is extracted in accordance with the human visual perception [5, 20-22]. It is apparent that HVS is a naïvely understood area so extraction of NSS features with ability to perfectly model the HVS is difficult [23]. A representative feature set to train a model for visual quality assessment is therefore a challenging problem [24].

Another approach to opinion un-aware IQA is learning based in which discriminative features are learned using dictionary-based [21, 25] or neural network based methods [24]. These features are believed to model the HVS better than the hand-crafted features. The problem with learning-based methods is that they require a large number of labelled training images to learn the peculiarities of the visual quality of images. Collection of a large number of subjectively scored images is difficult and therefore transfer learning and other such approaches are used to overcome this limitation. Transfer learning and dataset augmentation are among the approaches which are used to train CNN based networks [26].

In this work, we have targeted a feature-based approach to image quality assessment. We have explored the role of multi-scale [27] and multi-color space [28-30] in the extraction of features for image quality assessment. The statistics of the images extracted in different spatial resolution and color spaces is a unique introduction in the area of image quality assessment and it has provided reasonable improvement over state-of-the-art. A stacked ensemble of GPR is designed as a quality prediction model which is never explored for the problem of image quality assessment. Most of the existing approaches are based on SVM but GPR proved experimentally superior to SVM in generalization for image quality assessment. A comparison of other regression algorithms is provided to demonstrate the effectiveness of GPR in image quality assessment. The PIQI is tested on six benchmark databases and its performance is demonstrated in comparison to twelve state-of-the-art techniques. Cross-dataset evaluation experiments are performed to demonstrate the generalization of the proposed technique. The PIQI has provided competitive performance to most of the existing approaches.

The rest of the paper is organized as follows: Section II discusses the related work. Section III discuss the methodology, explaining the algorithmic design decisions including feature extraction, feature selection, choice of suitable regression algorithm and construction of stacked regression ensemble. Section IV provides the results and discussion. It explains the six benchmark databases and performance of PIQI in cross-dataset evaluation and comparison with state-of-the-art schemes. The paper is concluded in Section VI.

## 2. Related Work

No-reference image quality assessment is an important area of research due to its numerous applications. Different feature based and learning based techniques are proposed in the literature to address this problem. Most of these techniques are based on natural scene statistics or learning-based techniques. The learning-based techniques have gained popularity due to the availability of methods to train deep networks with a limited number of training images. However, no-reference quality assessment is a tough and challenging task due to a naïve understanding of the human visual system and variety of factors affecting the perceptual quality of digital images. Twelve significant no-reference image quality assessment methods are discussed in this section and are used for performance comparison with PIQI.

BIQI is a blind image quality index proposed by Moorthy et al. [14] which follows a two-step approach. As image quality assessment methods initially focused on distortion specific image quality assessment. The general-purpose application of image quality assessment doesn't specify the type of distortion and therefore distortion-specific methods are not suitable for such applications. Their two-step approach trained a distortion type classifier which identifies the type of distortion present in the image and then extracts the distortion specific features and calculates the image quality. Support Vector Machines (SVM) is used for identification of distortion type and Support Vector Regression (SVR) trained with wavelet coefficients modeled with generalized Gaussian distribution to estimate the quality of the image.

DIIVINE is a Natural Scene Statistics (NSS) based approach to perceptual quality assessment proposed by Moorthy et al. [15]. It is an extension of their previous work BIQI and uses a similar modular approach to first identify the type of distortion and then calculate distortion specific features for image quality assessment. The NSS based features are extracted in Discrete Wavelet Transform (DWT) domain using steerable pyramid decomposition. Generalized Gaussian distribution is fitted to get the coefficients as a feature and a final feature set of 88 features is extracted and SVM is trained for distortion type identification and SVR is trained for image quality assessment. DIIVINE has provided superior performance in comparison to Peak Signal-to-Noise Ratio (PSNR) and Structural Similarity Index Metric (SSIM) which are full-reference methods.

BLIINDS-II is an NSS based probabilistic model proposed by Saad et al. [31]. They model the DCT coefficients using generalized Gaussian distribution and then follow two approaches for quality prediction. It was demonstrated experimentally that score prediction using probabilistic modeling is better than training an SVR. The resulting model requires less training and use simpler modeling. The performance of the model is compared with two fell-reference and two no-reference models and it has provided competitive performance.

BRISQUE is a distortion generic NSS based image quality evaluator proposed by Mittal et al. [32]. This is a spatial domain approach and employee no transformation in the process of feature calculation. It calculates normalized luminance coefficients which occur in natural images and change in the presence of distortion. An SVR is trained on these features extracted on two scales based on the premises that distortion in natural images is scene specific. The model has performed well and shown superior correlation with MOS in comparison to all the available full-reference and no-reference methods. The feature set extracted is tried for distortion type identification and the idea of a distortion specific noise removal model is proposed.

NIQE is an NSS based opinion-unaware, distortion-unaware natural image quality evaluator proposed by Mittal et al. [33]. This approach uses similar features as in BRISQUE with only difference of selective feature extraction from non-overlapping patches. Unlike BRISQUE, the features are directly provided to multivariate Gaussian distribution model and the quality is predicted based on the deviation of the model parameters from the natural image. As there is no training involved so the approach is opinion-unaware as it didn't require subjective MOS. This approach has provided comparable performance with the state-of-the-art IQA approaches.

GM-LoG is a gradient magnitude and Laplacian of Gaussian-based image quality evaluator proposed by Xue et al. [10]. Local contrast of the image patches is extracted using normalized gradient magnitude and Laplacian of Gaussian. An SVR is trained using these features to construct the final model which is trained and tested on three benchmark databases and compared with five state-of-the-art no-reference approaches and three full-reference approaches. The proposed approach achieved the highest performance than compared approaches.

CORNIA is a dictionary learning-based approach to image quality proposed by Ye et al. [34]. The approach is an unsupervised feature learning approach which extracts local image patches to learn dictionary representation from unlabelled images. Effective image representations are obtained by soft assignment coding with max pooling. The dictionary is learned using k-mean clustering and max-pooling and SVR is used for training the quality assessment module. The algorithm is computationally efficient and its unsupervised learning-based approach makes it feasible for applications in different domains. The approach has been shown to outperform the state-of-the-art full-reference and no-reference approaches.

IL-NIQE is an opinion-unaware image quality evaluator similar to NIQE. It is proposed by Mittal et al. [11] and it requires no subjective score. This method uses a rich set of NSS including statistics of normalized luminance, statistics of mean subtracted and contrast normalized products, gradient and color statistics, and statistics of Log Gabor filter responses. These features are extracted from a set of pristine images without any subjective evaluation and multivariate Gaussian model is constructed. The quality scoring is done by calculating Bhattacharyya distance between these learned representations and individual image patches and the final score is provided by average pooling. The technique provided comparable performance with state-of-the-art techniques.

HOSA is an image quality assessment technique based on high order statistics aggregation proposed by Xu et al. [35]. Their work is based on the premises that feature-based approaches are not useful for images to have computer generated figures and sometimes generalize poorly and learning based approaches such as codebook learning require large sized codebooks which are difficult in a practical scenario. They have opted extraction and aggregation of mean, variance and skewness. The image is divided into non-overlapping patches and a codebook with 100 codewords is constructed using k-mean clustering. Each of local feature is assigned to the nearest cluster and difference of high order statistics is aggregated to build the global quality aware image representation. An SVR is trained on these image representations to obtain image quality score. Their approach has provided superior performance to the state-of-the-art approaches.

dipIQ is an opinion-unaware no-reference image quality evaluator proposed by Ma et al. [36]. The work is based on premises that there is a lack of reliable training database in comparison to the size of the image due to difficulty in conducting a subjective evaluation experiment. They have used RankNet which is a pairwise learning-to-rank algorithm to generate discriminable image pair quality index. The training experiment is not based on subjectively evaluated images but trained on scores obtained from FR methods. The approach is tested on four benchmark databases and has provided superior performance to the state-of-the-art approaches.

HaarPSI is a Haar wavelet decomposition based perceptual similarity index proposed by Reisenhofer et al. [37]. The image is decomposed using Haar wavelet and its coefficient are used as a features. The model is principally inspired by FSIM [29] which is a FR IQA method. The extracted features are used to compare local similarities between two images and identify the relative importance of these regions. Four benchmark databases are used to evaluate the correlation of the predicted score with subjectively evaluated scores. The comparison of dipIQ with state-of-the-art methods and its simpler structure, as well as less computational requirements, make it a good technique.

DeepBIQ is deep learning based blind image quality assessment approach which is proposed by Bianco et al. [26]. They have highlighted the role of deep learning-based approaches for image quality assessment. It was mentioned that deep learning-based approaches either extract features using pretrained Convolutional Neural Networks (CNN) trained for object recognition tasks or extracting features for specifically designing a network for image quality. Their approach divides the image in non-overlapping patches and calculating local image quality using CNN. The final quality score is computed through average pooling. They have tested their approach on a naturally distorted database of images and four synthetically distorted databases. The performance is compared with twelve state-of-the-art approaches and comparable performance is achieved.

The proposed approach has improved the feature set by extracting gradient statistics, statistics of normalized luminance and mean-subtracted contrast normalized products at three different spatial resolutions and multiple color spaces. This feature set has improved representational power for image quality. The introduction of a Gaussian process-based regression ensemble is another step to further improve the generalization performance as GPR is never explored for quality assessment task before and it has provided superior performance to commonly used SVM. This approach highlights the importance of features extracted at different spatial resolutions and different transform domain representation and such feature set can provide better quality assessment performance.

## 3. Methodology

It is observed that Natural Scene Statistics (NSS) is an excellent indicator of the perceptual quality of natural images [12, 14, 15, 31-33, 38, 39]. As a result, NSS based models are being widely used in Image Quality Assessment (IQA). Gradient statistics in combination with other natural scene statistics can be used in an optimal fashion to design an IQA system which can provide a good perceptual quality prediction. It is demonstrated in previous studies that distortion in image quality can be considered from features of local structural contrast, gradient, luminance and colour statistics [10, 28, 29]. In our study, we have combined an appropriate set of gradient statistics with Laplacian features and normalized luminance statistics. Moreover, we have presented the idea to extract features in different colour spaces as well as different spatial resolutions. The method discussed is a No-reference, opinion aware regression model which maps the features to mean opinion score acquired from subjective quality scores. The study is focused on the estimation of the quality of digital images which are distorted with commonly occurring distortions in digital images. The general flow of the proposed approach is provided in Figure 1.

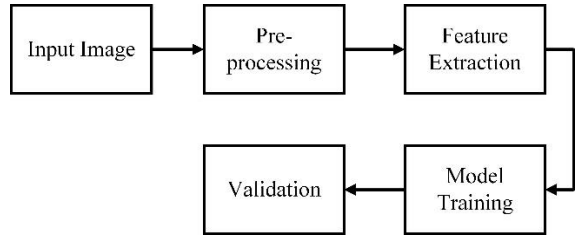

Figure 1. General block diagram of PIQI

The block diagram of the proposed scheme is provided in Figure 1. The input image is pre-processed for each feature extraction stage such as scaling and color space transformation. The second stage is the feature extraction stage depicted in Figure 2, which extracts gradient statistics, statistics of standardized luminance and mean-subtracted contras normalized products. These features are consolidated and provided to model training stage which learns an ensemble of Gaussian process regression as depicted in Figure 3. The final stage is model validation which use the unseen data to validate the model performance.

*3.1. Gradient Statistics*

The image gradient delivers extensive information regarding the structural properties of the image so it is extensively used in IQA research. Different gradient based statistics are used by various researchers and their approaches have provided varying performances [10, 12, 15, 28, 29, 40-42]. However, most gradient-based approaches are either full-reference (FR) or reduced-reference (RR) but some no-reference (NR) IQA approaches are also available which are mostly distortion specific with few exceptions [10-12, 43]. Owing to the success of these approaches in predicting blind image quality using gradient statistics there seems to be a strong bond between image quality and its gradient. We have relied on the gradient statistics extraction model of [12] which calculates Gradient Magnitude (GM), Relative gradient Orientation (RO), and Relative gradient Magnitude (RM). The estimated image gradient can be computed from Equation 1.

$$|\nabla I(i,j)| = \sqrt{I_x^2(i,j) + I_y^2(i,j)}, \qquad 1$$

Where $I_x$ and $I_y$ are values of directional derivatives in the horizontal $x$ and vertical $y$ directions corresponding to sample directions $i$ and $j$.

The distortions in image put a deep influence on gradient magnitude particularly when the image undergo blurring distortion [12]. Whereas in other types of distortions, the change in magnitude may not be much evident and though the image is still visually different from the original image. As an example, JPEG compression produces significant distortion which is visually apparent and still the gradient magnitude histograms of the perfect and distorted image are not significantly different making gradient magnitude alone less valuable feature. The gradient magnitude delivers information regarding brightness variations only whereas the human visual cortex is highly sensitive to orientation information as well [12]. As the distortions in image result in the modification of local image anisotropies, the use of gradient orientation along with gradient magnitude will be beneficial in the prediction of image quality. Similarly, relative gradient magnitude delivers much more information about the local structures in the image. The orientation of gradient delivers information additional to gradient magnitude which results in improved estimation performance [12]. The estimated gradient orientation can be given by Equation 2.

$$< \nabla I(i,j) = arc \tan \left( \frac{I_y(i,j)}{I_x(i,j)} \right), \qquad 2$$

The orientation of gradient can be measured in an absolute manner against the reference frame of the spatial coordinate system of the image or it could be measured relatively against the average image values.

The later model is more relevant to IQA as relative gradient orientation captures local degradations of image structure. Accordingly, three gradient maps, GM, RM, and RO are computed from $I_x$ and $I_y$ to describe the image gradient behavior with respect to perceptive quality over blocks of $M \times N$ dimensions. The GM is calculated using equation 1 whereas the RO can be calculated using Equation 3.

$$<\nabla I(i,j)_{RO} = <\nabla I(i,j) - <\nabla I(i,j)_{AVE},\qquad 3$$

And the average local orientation can be computed using Equation 4.

$$<\nabla I(i,j)_{AVE} = \arctan\left(\frac{I_y(i,j)_{AVE}}{I_x(i,j)_{AVE}}\right),\qquad 4$$

Where

$$I_x(i,j)_{AVE} = \frac{1}{MN}\sum_{(m,n)\in W}\sum I_x(i-m,j-n),$$

and

$$I_y(i,j)_{AVE} = \frac{1}{MN}\sum_{(m,n)\in W}\sum I_y(i-m,j-n),$$

Where $W$ defines a local neighborhood over which the derivative is taken. In this implementation a neighborhood of $3 \times 3$ is taken having $M = N = 3$, for which values of $W$ coordinates are $(-1,-1), (-1,1), (1,-1), (-1,0), (0,-1), (0,0), (0,1), (1,0)$ and $(1,1)$. Similarly, RM is given by Equation 5:

$$|\nabla I(i,j)|_{RM} = \sqrt{(I_x(i,j) - I_x(i,j)_{AVE})^2 + (I_y(i,j) - I_y(i,j)_{AVE})^2},\qquad 5$$

The directional gradient components $I_x$ and $I_y$ are computed using Gaussian partial derivative filters. In comparison to other filters such as Prewitt and Sobel, the Gaussian partial derivative can be treated as a smoothed edge filter which is beneficial for computing RO and RM. Equation 6 can be used for computation of Gaussian partial derivative:

$$\nabla_\gamma G(x,y,\sigma) = -\frac{\gamma}{2\pi\sigma^4}\exp\left(-\frac{x^2+y^2}{2\sigma^2}\right),\qquad 6$$

As mentioned in [12] variations in statistical distributions of gradient and relative gradient measures due to distortions can be used for quantification of perceived distortion in images. These gradient distributions are characterized by using a histogram of variance as described in [44]. Though all the characteristics of histograms cannot be expressed through variances, still they signify the most notable characteristics. The variance of a histogram $h(x)$ can be defined by Equation 7:

$$Var(h) = \sum_x\left(h(x) - \bar{h}\right)^2,\qquad 7$$

Resultantly, we obtain a three-dimensional feature vector:

$$Feat\_Vec = [V_{GM}\quad V_{RO}\quad V_{RM}],$$

The three-dimensional feature vector is usually calculated for grayscale image [12] which is not sufficient to capture the change in gradient statistics. We have calculated gradient statistics for red, green and blue channels of RGB image; luminance, blue chrominance and red chrominance of YCbCr image, and hue, saturation and intensity of HSI image. The extended feature set was observed to provide better prediction than the grayscale image only. Moreover, natural images exhibit multi-scale characteristics for distortion and visual perception [31, 45], which requires representation. The features are extracted at three different scales by downsampling image array into $\frac{1}{2}$ and $\frac{1}{4}$ and resulting in a three-fold increase in the number of features as depicted in Figure 2. It is to note that further scaling to extract more feature didn't result in a significant gain in predictive accuracy.

*3.2. Statistics of Standardized Luminance*

It is discussed in [44] that the standardized luminance of an image $I$ adapt to a Gaussian distribution. The standardization of the natural image can be done by using equation 8.

$$\bar{I}(i,j) = \frac{I(i,j) - \mu(i,j)}{\sigma(i,j)+1},\qquad 8$$

Whereas i and j are spatial coordinates of pixels and

$$\mu(i,j) = \sum_{k=-K}^{K}\sum_{l=-L}^{L} \omega_{k,l}\, I(i+k, j+l),$$

$$\sigma(i,j) = \sqrt{\sum_{k=-K}^{K}\sum_{l=-L}^{L} \omega_{k,l}\, [I(i+k, j+l) - \mu(i,j)]^2},$$

are the mean and contrast of the local image. The structural distortion is characterized by local MSCN coefficients and from products of adjacent pair distribution of these coefficients. These product coefficients $I(i,j)$ follow a unit Gaussian distribution on digital images which are not perceptually distorted [44]. This Gaussian model is disrupted when the images undergo some distortion. Measurement of deviance of $I(i,j)$ from the Gaussian model indicate the level of distortion. According to suggestion in [32] & [33], a zero-mean Generalized Gaussian Distribution (GGD) model can generally channelize the dispersal in the incidence of distortions. The density function of GGD can be given by Equation 9.

$$g(x; \alpha, \beta) = \frac{\alpha}{2\beta\Gamma\left(\frac{1}{\alpha}\right)} exp\left(-\left(\frac{|x|}{\beta}\right)^{\alpha}\right),$$

Where $\Gamma(.)$ is the gamma function

$$\Gamma(x) = \int_{0}^{\infty} t^{x-1} e^{-t} dt, \ x > 0,$$

The parameters $\alpha$ and $\beta$ in equation 9 are effective features which can be estimated using the moment matching approach with reliance [46]. These two features are extracted on three different scales to capture the multi-scale characteristics of distortion and visual perception [31, 45]. Moreover, the extraction of features is not performed in grayscale but the features are extracted in three colour spaces and three different spatial resolutions as depicted in Figure 2.

*3.3. Statistics of MSCN Products*

Image quality information can also be captured through products of adjacent pairs of MSCN coefficients as described in [33] and [32]. The distribution of product pairs particularly $I(i,j)\, I(i,j+1), I(i,j)\, I(i+1, j+1)$, $I(i,j)\, I(i+1, j-1)$ and $I(i,j)I(i+1,j)$ is calculated to describe image quality. These products are demonstrated on perfect and distorted images following a zero mode asymmetric GGD (AGGD) [47]. AGGD can be calculated using Equation 10.

$$g_{\alpha}(x; \gamma, \beta_l, \beta_r) = \begin{cases} \frac{\gamma}{(\beta_l + \beta_r)\Gamma\left(\frac{1}{\gamma}\right)} exp\left(-\left(\frac{-x}{\beta_l}\right)^{\gamma}\right), & \forall_x \leq 0 \\ \frac{\gamma}{(\beta_l + \beta_r)\Gamma\left(\frac{1}{\gamma}\right)} exp\left(-\left(\frac{x}{\beta_r}\right)^{\gamma}\right), & \forall_x > 0 \end{cases},$$

The mean of the AGGD is given by Equation 11:

$$\eta = (\beta_r - \beta_l)\frac{\Gamma\left(\frac{2}{\gamma}\right)}{\Gamma\left(\frac{1}{\gamma}\right)},$$

These four parameters ($\gamma, \beta_l, \beta_r, \eta$) are also effective features which can describe image quality and are calculated in four orientations. These features are calculated on three different scales in order to capture the multi-scale characteristics of distortion and visual perception [31, 45].

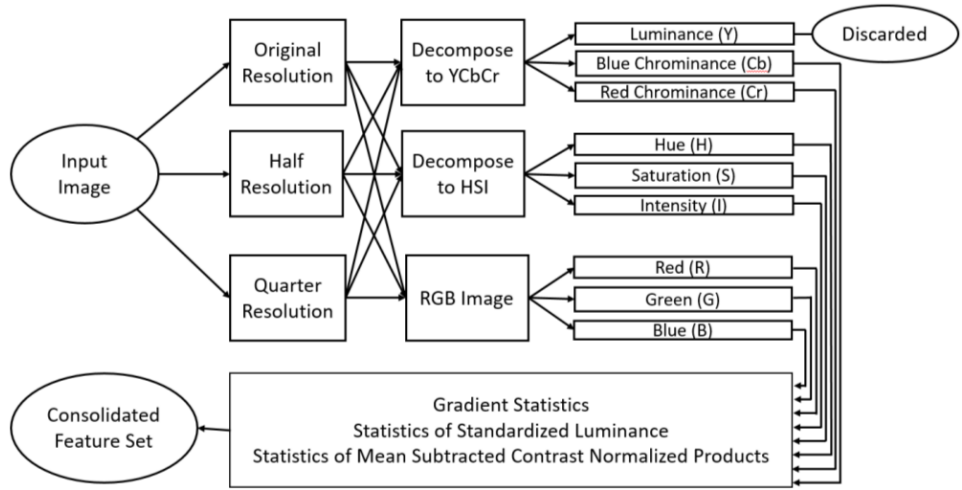

Figure 2. Flow chart of feature extraction and consolidation

The final feature set is formed by appending the three feature subsets into the consolidated feature set containing 192 features.

*3.4. Selection of Regression Algorithm*

Linear and non-linear are the two major classes of regression algorithm. Linear regression algorithms such as Multiple Linear Regression (MLR), a parametric method, try to estimate the parameters of the linear equation from data to model the prediction of targets. Such models require a small amount of data and provide good baseline performance for most of the cases. Non-linear regression algorithms, on the other hand, are used to model the complex relationship between the predictors and response variable. In this work, we have mainly focused on non-parametric regression algorithms which require supervised data to learn the predictor-response relationship. Table 1 provides the RMSE and R-squared for different regression algorithms and it is evident from the result that Gaussian Process Regression (GPR) has provided consistently good performance and GPR with Matern 5/2 kernel has given lowest RMSE of 0.48378 and highest value for the coefficient of determination (R-squared). On the basis of Table 1, GPR is most suitable regression algorithm for the problem at hand.

Table 1. Performance of the regression algorithms

| Regression Algorithm | RMSE | R2 |
| --- | --- | --- |
| MLR | 0.76473 | 0.62 |
| SVM (Linear) | 0.79207 | 0.59 |
| SVM (Quadratic) | 0.93812 | 0.43 |
| SVM (Gaussian) | 0.63547 | 0.74 |
| GPR (Sq. Exponential) | 0.49728 | 0.84 |
| GPR (Matern 5/2) | 0.48360 | 0.85 |
| GPR (Matern 3/2) | 0.48392 | 0.85 |
| GPR (Exponential) | 0.48418 | 0.85 |
| GPR (Rational Quadratic) | 0.48395 | 0.85 |
| Decision Tree | 0.85636 | 0.55 |
| ANN | 0.52147 | 0.79 |

*3.5. The ensemble of Gaussian Process Regression*

The performance of some regression algorithms in Table 1 shows that GPR is superior in comparison to other approaches. The performance and generalization of the model could be further improved by using an ensemble of GPR. We have adopted bagging of GPRs with subsets of training samples and feature space to construct an ensemble. Increase in the number of features may result in reduced variance and increased bias, conversely the reduction in the number of features may result in decreased bias and increased variance. Bagging of weak learners with a subset of feature space, on the other hand, will attempt to reduce variance while keeping the bias same.

The dataset is divided into 70% data for training and 15% for validation and 15% for testing. The testing data is only used to measure the performance of the model after completion of the model building process. GPR is constructed with "Mattern 5/2" kernel and its hyper-parameters are tuned using Bayesian Regularization (BR). The model is trained by random sub-sampling of training data and a hundred models are trained initially. Combining all the trained model through averaging or weighted averaging is one way of ensembling but it is a naive approach and can be improved further. A good ensemble is the one having diverse and accurate base learners, we have introduced diversity by random subsampling and feature subspace selection but accuracy remains a concern. It is hypothesized that combining many base learners will yield better than combining all the base learners. Therefore, Stepwise Linear Regression (SLR) as a second step meta-learner will perform weighted selection of some base learners. The final model consists of 27 base learners which are chosen and assigned weights through SLR as meta-learner.

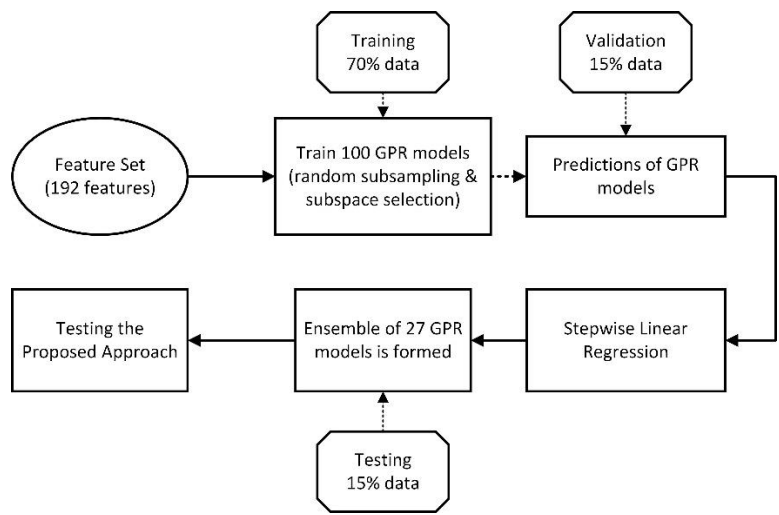

Figure. 3. Flow Chart of Ensemble Learning

The flow chart of Figure 3 describes the steps followed for ensemble learning. The dataset is initially divided into three trains, validation, and test split. The training data is randomly subsampled with feature subset selection and a GPR model with fine-tuning is trained for each iteration. Training iterations were repeated for 100 times and GPR models were generated which were provided to SLR and the validation data is used to judge the performance of the SLR model. The final model consisted of 27 GPR models which were selected using SLR having a coefficient acting as weight in this case. The final ensemble model is evaluated on 15% testing data which was never exposed during the model building process.

*3.6. Convergence of Ensemble Learning Algorithm*

Stacked ensembles is a class of learning algorithms which perform learning on predictions of other learning algorithm. In our scenario, we have trained linear regression on predictions of GPR models with stepwise addition of each learner. Stepwise addition is opted based on assumption that "many could be better then all" [48]. The meta-learning algorithm (linear regression) is performed by sequentially adding more GPR models into final model until the improvement in performance is halted. Our final model saturated after addition of 27 GPR models and started decreasing after inclusion of nearly 30 GPR models. The graph of Figure 4 provides the change in RMSE value with sequential addition of GPR models in the final model. The final model has converged at 27 GPR models and inclusion of further models resulted no-improvement or decreased performance in terms of RMSE.

## 4. Results & Discussion

There are two types of IQA benchmark databases: naturally distorted and synthetically distorted. The first category contains the databases which are distorted during the process of image acquisition such as camera shake, low light, noise, etc. The second categories of databases are the one which contains images with simulated distortions. This type of databases contains pristine images which are processed with distortion models causing single or multiple distortions in images. These types of databases are most common and are also used in full-reference and reduced-reference image quality assessment. We have evaluated our model on six databases among which one is naturally distorted and others are synthetically distorted.

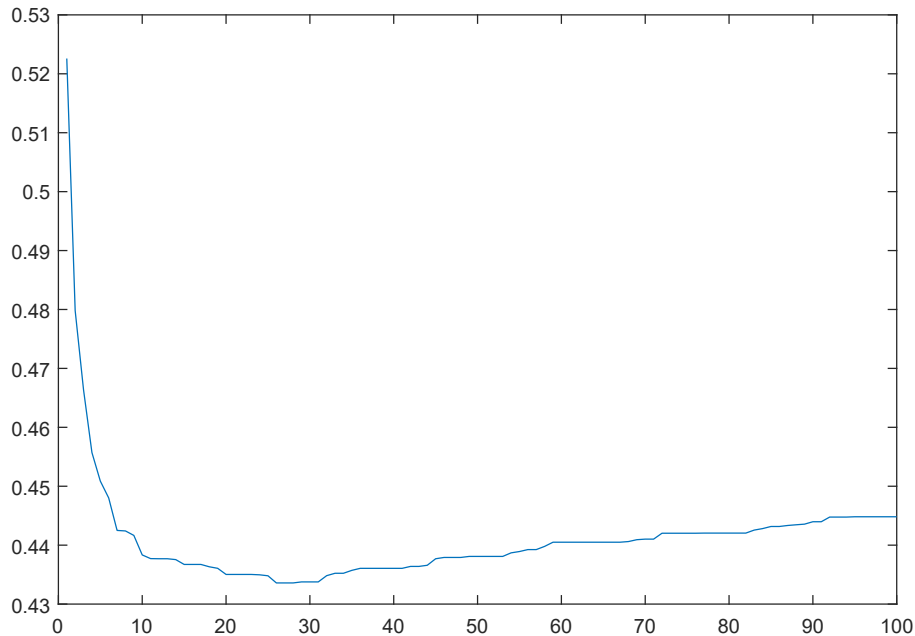

Figure 4 RMSE vs Number of included GPR models

Table 1. List of databases

| Database Name | No of reference images | No of distorted images | Type | Scoring Method | Range |
| --- | --- | --- | --- | --- | --- |
| LIVE-I | 29 | 460 | synthetic | DMOS | 0-100 |
| LIVE-II | 29 | 982 | synthetic | DMOS | 0-100 |
| TID2008 | 25 | 1700 | synthetic | MOS | 1-10 |
| TID2013 | 25 | 3000 | synthetic | MOS | 1-10 |
| CSIQ | 30 | 900 | synthetic | DMOS | 0-1 |
| Toyama | 24 | 196 | synthetic | DMOS | 1-5 |
| Live in the Wild | - | 1169 | Natural | DMOS | 0-100 |

Table 2 provides the list of seven databases which are used for experimentation. It is to be noted that LIVE release I and II use similar scoring mechanism and their scores are translated in 0-100 range so the model trained on one of these two databases can be combined or cross-validated on the other. Similarly, TID2008 and TID2013 databases have similar scoring with 1-10 range. These datasets have been evaluated by humans from different countries. These datasets also have similar scoring so they can also be combined and used for cross-validation performance of each other. CSIQ and Toyama databases have synthetic distortions and DMOS is used as a scoring mechanism. Live in the Wild is the dataset which have natural distortions with no-reference images. The range of all these databases is compressed to 0-1 to provide a unified comparison.

The performance of the PIQI is tested on the basis of the correlation between human subjective evaluation and estimation of the score through the proposed index. It is to note that the coefficient of determination (R2) and Root Mean Squared Error (RMSE) are also computed to check the fit of the model and error. Pearson Linear Correlation Coefficient (PLCC) measures the linear correlation between the subjective evaluation and estimated scores. Spearman Rank Order Correlation Coefficient (SROCC) and Kendall Rank Order Correlation Coefficient (KROCC) measures the degree of association between subjective evaluation and estimated scores. It measures the similarity of the ranked ordering of the data and gives a measure of how well the relationship between the two scores can be described through a monotonic function. Mostly, SROCC provided slightly higher value then KROCC whereas similar inference can be made from any of these.

Table 2. Performance of PIQI on benchmark databases

| Databases | R2 | RMSE | PLCC | SROCC | KROCC |
|---|---|---|---|---|---|
| TID2013 | 0.8780 | 0.4300 | 0.9685 | 0.9717 | 0.8636 |
| TID2008 | 0.8590 | 0.4316 | 0.9487 | 0.9504 | 0.8161 |
| CSIQ | 0.9480 | 0.0552 | 0.9802 | 0.9776 | 0.8696 |
| LIVE-I | 0.9650 | 4.5840 | 0.9779 | 0.9554 | 0.8267 |
| LIVE-II | 0.9380 | 3.5696 | 0.9752 | 0.9741 | 0.8597 |
| Toyama | 0.9840 | 0.2960 | 0.9920 | 0.9857 | 0.9136 |
| Wild | 0.7420 | 10.2912 | 0.8617 | 0.8376 | 0.6535 |

Table 2 indicates that there is a very good correlation between human subjective evaluation and the estimated scores. It is to be noted that correlation is comparatively low in case of Live in the Wild which is a database with natural distortions (multiple distortions occurring naturally) and human evaluations are obtained using single stimulus only. This database contains larger variabilities in the form, intensity, and mix of distortions and the number of training images are not representative and is harder to capture. Moreover, humans didn't have a reference and their opinion could be deeply biased based on the content. More efforts such as size of the training database and images having different type of contents are required to improve the quality score estimation of naturally distorted images.

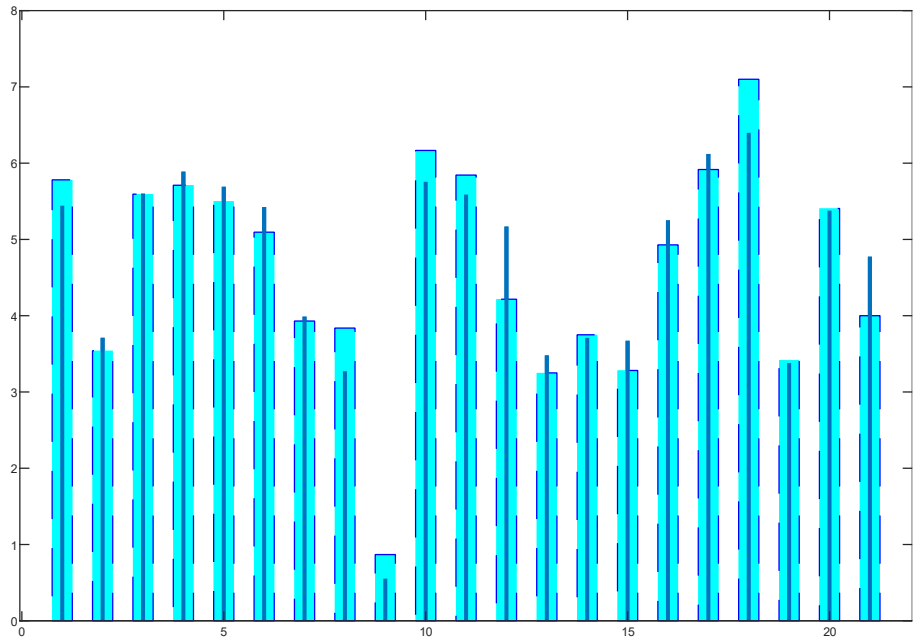

Figure 5 Ground Truth (cyan) and Predicted (blue) bar chart for 20 samples.

Figure 5 provides the bar chart of ground truth (cyan) values of quality score (MOS) and predicted values of quality score for TID2013 database. Similar pattern can be observed for other experiments and the whole database but it is shown for 20 samples of TID2013 database for the sake of conciseness.

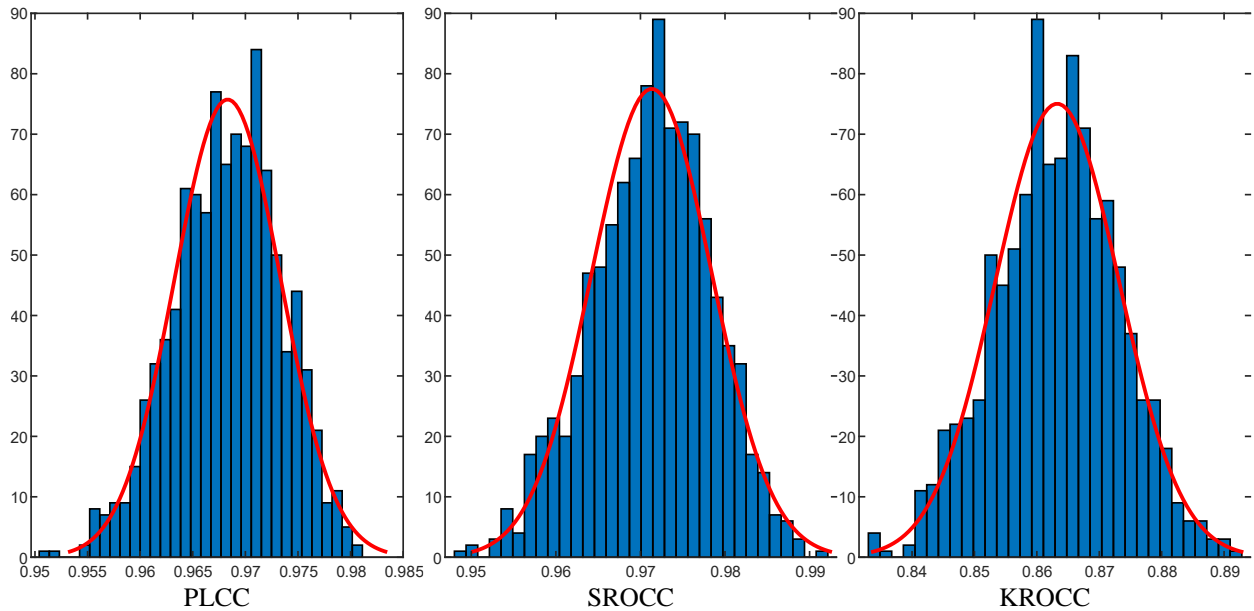

Figure 6 Histograms with Gaussian fitting for 1000 random train-test split for Pearson, Spearman and Kendall's correlation coefficients on TID2013 image database.

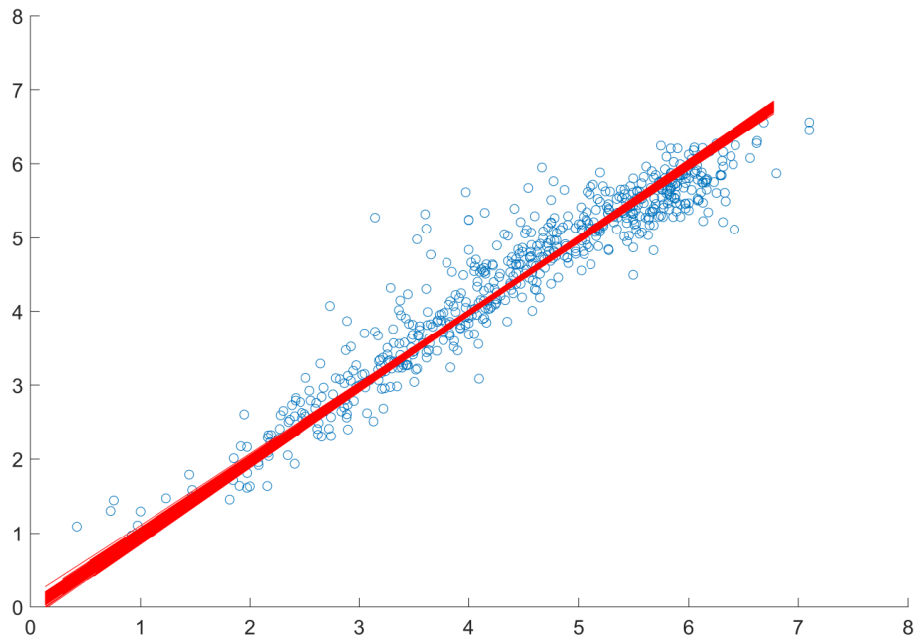

Figure 7 Scatter plot between predicted (median value among 1000 iterations) and ground truth scores for TID2013 database along with linear fitting of 1000 random train-test splits

The scores of Tables 2 are the median values of 1000 iterations of train-test splits. As the benchmark databases are not provided with a train-test splits and the codes of all the comparing algorithms are not available open source, making it is difficult to make a fair comparison. We have therefore followed a practice to repeat the experiment for 1000 times in order to make a fair comparison. Figure 6 provides the histograms of PLCC, SROCC and KROCC values for 1000 experiments using TID2013 database. Similar results are obtained for other databases but are not reported in order to follow conciseness. It is to be noted that the distribution of these values follows Gaussian and most of the values are concentrated near the median values with very few values spreading away from the median value. Moreover, the spread provides a way to identify that the model has provide a good fit which doesn't vary largely with the change in train-test data split and it can demonstrate a good generalization. Similarly, Figure 7 has provided a scatter plot between the predictions of the experiment with median PLCC values and the ground truth values. Linear fitting is provided in red color for predictions of 1000 iterations indicating that the fitting doesn't diverge much with a change in train-test data split. Figure 8 provides the scatter plots between the median predicted and ground truth values for the six-image database along with the linear fitting line in red color. It can be noted from these plots, that the proposed approach is robust and provide consistent performance on different databases.

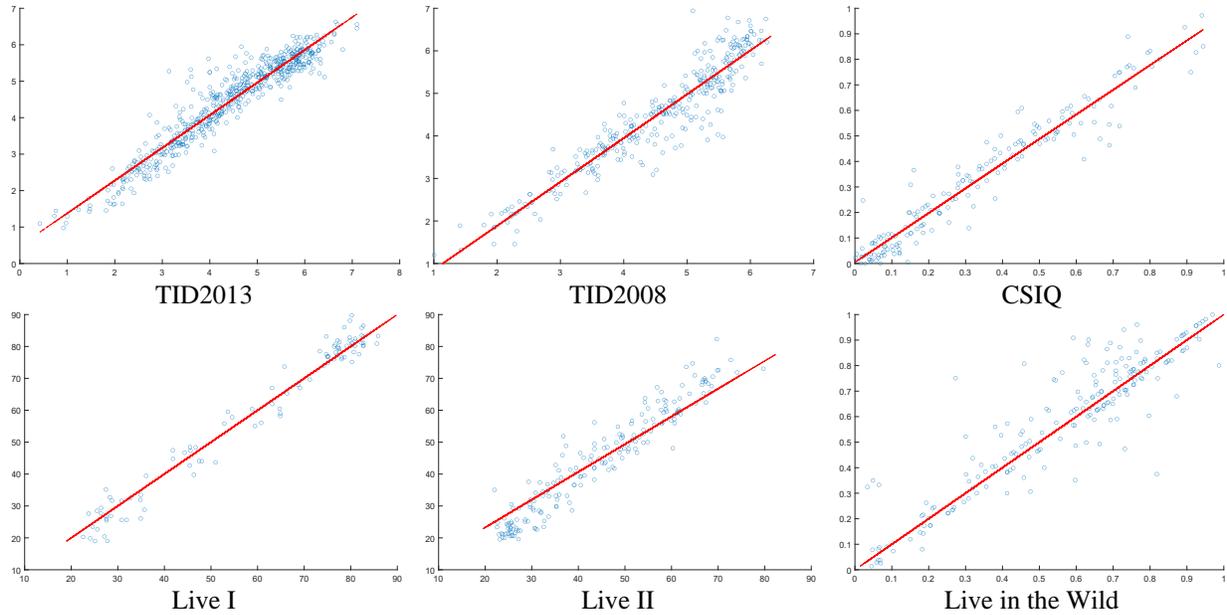

Figure 8 Scatter plot between the median predicted and the ground truth values along with the linear fitting line

The residuals play an important role in identification of model's behaviour. The histogram of residuals is used to check if the variance follows a Gaussian distribution. A symmetrically distributed normal histogram which is evenly distributed around zero indicates that the assumption of the normality is true and the model's underlying assumptions are true. The same is checked in the probability plot for normality and shown in Figure 9 which provides a histogram of residuals with Gaussian fitting and probability plot for normal distribution of residuals. It is evident from the Figure 9 that the residuals likely follow the assumptions of normality.

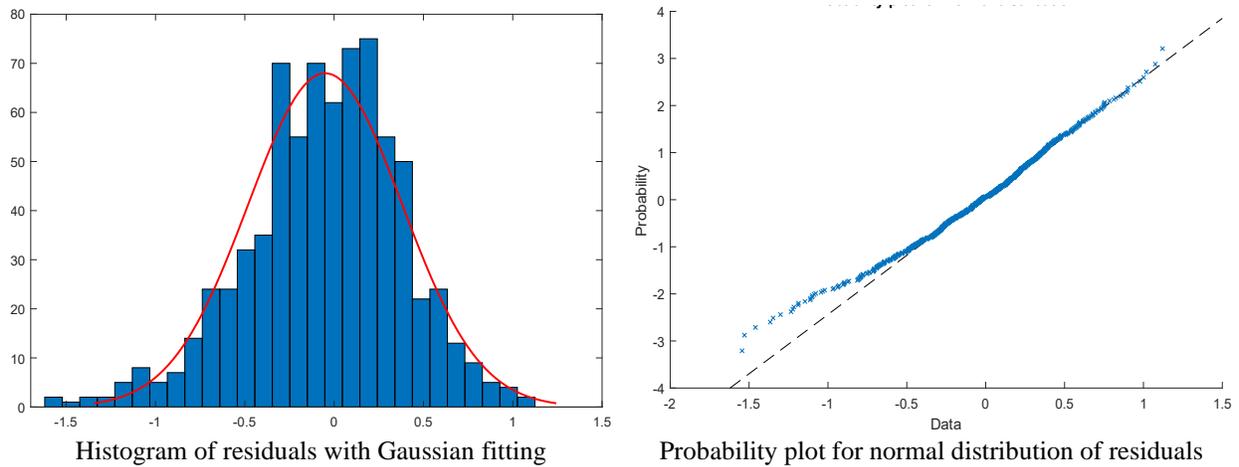

Histogram of residuals with Gaussian fitting    Probability plot for normal distribution of residuals

Figure 9 Normality testing of residuals for TID2013 database

The residual plots of Figure 9 are shown for one iteration only having median value of PLCC. A box plot of residuals for 20 iterations on TID2013 database is shown in Figure 10 to give an idea of distribution of residuals. It can be noted that mean of most of the distributions is zero or very close to zero with some outliers spanning up to 1.83.

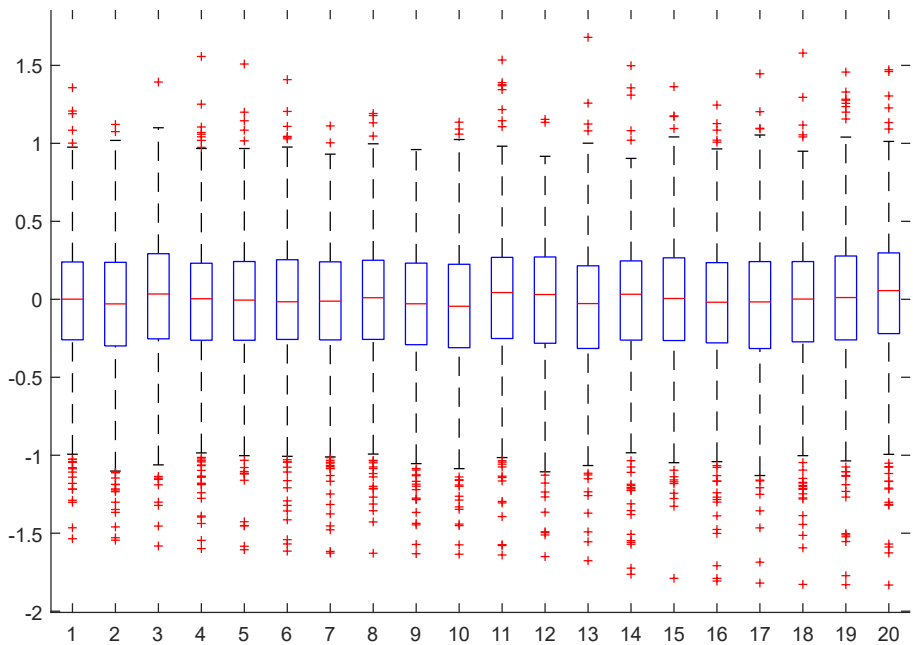

Figure 10 Box plot of residuals for 20 iterations

### 4.1. Cross-Dataset Evaluation

No-reference quality assessment is a challenging task when it comes to generalization. The trained model should perform similarly for new images as it is performing on the dataset which is used for training. The limitations in training of a successful IQA model comes from the complex nature of problem which involves variable sources of image distortions and subjective nature of scoring which varies from human to human. Human evaluation can be influenced due to socio-economic, educational and cultural background. The model can learn the peculiarities of an image database if it contains some specific number of distortions and result in over-fitting. A larger dataset having diverse distortion types and evaluated by non-expert humans of different cultural background can provide better generalization. TID2008 and TID2013 tried to follow these points and they have used more distortion categories and acquired human evaluation from different countries to represent the general behavior. Live in the Wild has natural distortions and they have used online crowdsourcing for their scoring and a total of 150,000 persons has rated 1169 images making 128 evaluators for each image on average. This dataset has diverse distortions occurred naturally but have smaller number of images and therefore models trained on this dataset doesn't provide very good performance.

For cross-dataset evaluation we have performed two experiments, the first one is trained on LIVE-II and tested on LIVE-I and trained on LIVE-I and tested on LIVE-II. The second is trained on TID2013 and tested on TID2008 and trained on TID2008 and tested on TID2013. The same evaluation cannot be performed for other datasets as they have a different scoring mechanism, the range of scores and experimental setup. In order to conduct the evaluation of other dataset session alignment and rescaling can be used. Table 4 provides the result of these experiments. It is to be noted that the results of LIVE-I in the first experiment are obtained by training the model on LIVE-II and vice versa. Similar testing methodology is performed for Experiment-II.

Table 3. Cross-dataset evaluation performance of the PIQI

| Evaluation Measure | Experiment-I | | Experiment-II | |
|---|---|---|---|---|
| | LIVE-I | LIVE-II | TID2008 | TID2013 |
| RMSE | 6.4516 | 3.5751 | 0.5245 | 0.8321 |
| PLCC | 0.9475 | 0.9322 | 0.941 | 0.8311 |
| SROCC | 0.9693 | 0.9477 | 0.9464 | 0.8207 |
| KROCC | 0.9187 | 0.8409 | 0.8044 | 0.6447 |

*4.2. Performance Comparison with Existing Approaches*

Table 5 provides a comparison of the PIQI with existing techniques. The comparison is made based on three parameters SROCC, PLCC and RMSE. It is to be noted that higher values of PLCC indicates good linear correlation of predicted quality score with human opinion and higher values of SROCC indicates good monotonicity. The PIQI is consistently high performing among five of the six benchmark databases. DeepBIQ is a deep learning-based model which has provided comparable performance in case of TID2008 database and better performance in case of Live in the Wild challenge database. It is to be noted that the first five databases contain simulated distortions and are called synthetic distortion database whereas the Live in the Wild challenge database contains distortions which are introduced during the process of image acquisition only. Moreover, the Live in the Wild database contains scoring through crowdsourcing using the single stimulus whereas the other databases contain quality scoring obtained using double stimulus which provides pristine and distorted images in comparison for scoring and therefore provides more reliable scores.

Table 4 Performance comparison of PIQI with existing schemes on six benchmark databases

| Database | | BIQI [14] | BLIINDS-II [31] | BRISQUE [32] | CORNIA-10K [49] | CORNIA-100 [49] | DeepBIQ [26] | DIIVINE [15] | dipIQ [36] | GM-LOG [10] | HaarPSI [37] | HOSA [35] | ILNIQE [11] | NIQE [33] | PIQI |
|---|---|---|---|---|---|---|---|---|---|---|---|---|---|---|---|
| LIVE | SROCC | 0.8642 | 0.9302 | 0.9409 | 0.9417 | 0.8572 | 0.98 | 0.9162 | 0.958 | 0.9503 | 0.9683 | 0.9504 | 0.902 | 0.9135 | 0.9741 |
| | PLCC | 0.8722 | 0.9357 | 0.945 | 0.9434 | 0.8579 | 0.97 | 0.9172 | 0.957 | 0.9539 | - | 0.9527 | 0.9085 | 0.9147 | 0.9752 |
| | RMSE | 13.285 | 9.6189 | 8.9048 | 9.0204 | 14.0175 | - | 10.810 | - | 8.1723 | - | 8.2858 | 11.4007 | - | 3.5696 |
| TID2008 | SROCC | 0.8438 | 0.8982 | 0.9357 | 0.899 | - | 0.95 | 0.893 | - | - | 0.9097 | 0.917 | - | 0.6503 | 0.9504 |
| | PLCC | 0.8704 | 0.9219 | 0.9391 | 0.9347 | - | 0.95 | 0.9039 | - | 0.9269 | - | 0.899 | - | 0.7409 | 0.9487 |
| | RMSE | 8.4704 | 1.1389 | 1.1329 | - | - | - | - | - | - | - | - | - | 1.144 | 0.4316 |
| TID2013 | SROCC | 0.8191 | 0.8786 | 0.8917 | 0.8998 | 0.8276 | 0.96 | 0.8753 | 0.894 | 0.9282 | 0.8732 | 0.9521 | 0.8871 | 0.317 | 0.9717 |
| | PLCC | 0.8407 | 0.9053 | 0.9176 | 0.9277 | 0.8554 | 0.96 | 0.8859 | 0.877 | 0.9439 | - | 0.9592 | 0.903 | 0.426 | 0.9685 |

|  |  |  |  |  |  |  |  |  |  |  |  |  |  |  |
|---|---|---|---|---|---|---|---|---|---|---|---|---|---|---|
|  | RMSE | 0.7569 | 0.5921 | 0.5534 | 0.5239 | 0.728 | - | 0.6474 | - | 0.4629 | - | 0.3941 | 0.602 | - | 0.4300 |
| CSIQ | SROCC | 0.8115 | 0.914 | 0.9099 | 0.893 | 0.8216 | 0.96 | 0.876 | 0.949 | 0.9228 | 0.9604 | 0.9298 | 0.8885 | 0.8829 | 0.9776 |
|  | PLCC | 0.8476 | 0.9323 | 0.9278 | 0.9175 | 0.8476 | 0.97 | 0.8983 | 0.93 | 0.9408 | - | 0.948 | 0.9173 | 0.9345 | 0.9802 |
|  | RMSE | 0.1491 | 0.101 | 0.1044 | 0.1123 | 0.1492 | - | 0.122 | - | 0.095 | - | 0.0887 | 0.1098 | 0.1089 | 0.0552 |
|  | SROCC | 0.5949 | 0.7995 | 0.85 | 0.8565 | 0.716 | - | 0.8198 | - | 0.8551 | - | 0.9066 | 0.7772 | - | 0.9857 |
| Toyama | PLCC | 0.5948 | 0.7672 | 0.8269 | 0.8434 | 0.7074 | - | 0.7915 | - | 0.8371 | - | 0.8999 | 0.7798 | - | 0.9920 |
|  | RMSE | 1.0039 | 0.7914 | 0.7099 | 0.6781 | 0.8906 | - | 0.772 | - | 0.6897 | - | 0.5409 | 0.7759 | - | 0.2960 |
|  | SROCC | 0.291 | 0.447 | 0.597 | - | - | 0.91 | 0.43 | - | - | - | 0.65 | - | 0.48 | 0.8376 |
| Live In the Wild | PLCC | 0.388 | 0.483 | 0.63 | - | - | 0.89 | 0.468 | - | - | - | - | - | 0.42 | 0.8617 |
|  | RMSE | - | - | - | - | - | - | - | - | - | - | - | - | - | 10.291 |

Empty cells indicate that no performance is mentioned in the original paper or the literature. The performance of the reported algorithms in Table 5 can be compared absolutely indicating that the proposed approach has provided superior performance to the most of existing approaches but a test of statistical significance can provide a better comparison.

**Statistical Significance Test**

Although the difference in correlation and RMSE between the predicted and MOS values is provided in Table 4 and can be used for absolute comparison. These scores are reported by performing 80/20 train/test splits for 1000 times and taking median value. The differences between the reported performances are visible but they may not be statistically significant. The evaluation of the statistical significance is performed by hypothesis testing using one-sided t-test and the results are reported in Table 5. The null hypothesis is stated as the mean SROCC of column algorithm is equal to the mean SROCC of the row algorithm. The alternate hypothesis states that the mean SROCC of the column algorithm is greater than the mean SROCC of the row algorithm. The null hypothesis is tested with 95% confidence interval. Therefore, a value of '1' in the Table 5 indicates that the column algorithm is statistically superior to the row algorithm while a value of '-1' indicates that the column algorithm is statistically worse than the row algorithm. Whereas a value of '0' indicates that the column and row algorithms are not statistically distinguishable.

Table 5 Results of One-sided T-Test with 95% confidence interval

|  | BIQI [14] | BLIINDS-II [31] | BRISQUE [32] | CORNIA-10K [49] | CORNIA-100 [49] | DeepBIQ [26] | DIIVINE [15] | dipIQ [36] | GM-LOG [10] | HaarPSI [37] | HOSA [35] | ILNIQE [11] | NIQE [33] | PIQI |
|---|---|---|---|---|---|---|---|---|---|---|---|---|---|---|
| BIQI [4] | 0 | 1 | 1 | 1 | -1 | 1 | 1 | 1 | 1 | 1 | 1 | 1 | -1 | 1 |
| BLIINDS-II [22] | 1 | 0 | 1 | -1 | 1 | 1 | -1 | -1 | 1 | -1 | 1 | -1 | 1 | 1 |
| BRISQUE [23] | 1 | 1 | 0 | -1 | 1 | 1 | 1 | -1 | 1 | -1 | 1 | 1 | 1 | 1 |
| CORNIA-10K [39] | 1 | -1 | -1 | 0 | 1 | 1 | 1 | -1 | 1 | -1 | 1 | 1 | 1 | 1 |
| CORNIA-100 [39] | -1 | 1 | 1 | 1 | 0 | 1 | 1 | 1 | 1 | -1 | 1 | 1 | -1 | 1 |
| DeepBIQ [17] | 1 | 1 | 1 | 1 | 1 | 0 | 1 | 1 | 1 | -1 | -1 | 1 | 1 | -1 |
| DIIVINE [5] | 1 | -1 | 1 | 1 | 1 | 1 | 0 | 1 | 1 | -1 | 1 | -1 | -1 | 1 |
| dipIQ [27] | 1 | -1 | -1 | -1 | 1 | 1 | 1 | 0 | -1 | -1 | -1 | -1 | -1 | 1 |
| GM-LOG [1] | 1 | 1 | 1 | 1 | 1 | 1 | 1 | -1 | 0 | -1 | -1 | 1 | -1 | 1 |
| HaarPSI [28] | 1 | -1 | -1 | -1 | -1 | -1 | -1 | -1 | -1 | 0 | -1 | -1 | -1 | -1 |
| HOSA [26] | 1 | 1 | 1 | 1 | 1 | -1 | 1 | -1 | -1 | -1 | 0 | 1 | 1 | 1 |
| ILNIQE [2] | 1 | -1 | 1 | 1 | 1 | 1 | -1 | -1 | 1 | -1 | 1 | 0 | -1 | 1 |
| NIQE [24] | -1 | 1 | 1 | 1 | -1 | 1 | -1 | -1 | -1 | -1 | 1 | -1 | 0 | 1 |

| PIQI | 1 | 1 | 1 | 1 | 1 | -1 | 1 | 1 | 1 | -1 | 1 | 1 | 1 | 0 |

A value of '1' indicates that the column algorithm is superior to the row algorithm whereas a value of '-1' indicates that the row algorithm is superior. The value of '0' indicates that the algorithms are indistinguishable or equivalent.

**Time Complexity**

The experiments are performed on Dell T5600 with dual Intel® Xeon® Processor E5-2687W with 512 GB SSD, 32 GB RAM and RTX 2070 GPU. The software includes MATLAB ® 2019a on Windows 10 Pro 64-bit. The time for feature extraction, GPR training, ensemble learning and complete training time for LIVE release 1 and testing time for single image are provided in the Table 6. It is to be noted that feature extraction task takes the most time due complexity of the feature extraction task but the reported time is for an unoptimized version of the code. An optimized code can provide lesser feature extraction time. The training of GPR is performed with hyperparameter search as it provides slight increase in the performance of the model at an expense of large increase in time but the training is a one-time process and it is not required at the runtime. Therefore, the GPR models are tuned for hyperparameters resulting in relative increase in the training time. On contrary the quality prediction task extract features for a single image and then perform score calculation, the time for quality prediction is mostly taken by the feature extraction task which is nearly three seconds whereas the other model parameters contribute minimally towards the time complexity.

Table 6 Time Complexity of the PIQI

| Task | Number of Images | Time (seconds) |
|---|---|---|
| Feature Extraction | 460 images | 1337 |
| Training of Single GPR model | 322 images | 1 |
| Training of GPR with hyperparameter tunning | 322 images | 114 |
| Training of 100 GPR models | 322 images | 8337 |
| Ensemble Learning with stepwise addition | 69 images | 113 |
| Total Training Time |  | 9787 |
| Quality Prediction | 1 image | 3 |

**5. Conclusion**

An efficient no-reference perceptual image quality index (PIQI) is proposed in this study which predicts the quality of a digital image using a stacked ensemble of Gaussian Process Regression. The training is done by extracting features from training images and providing MOS/DMOS as response variable which is obtained through subjective evaluation by human observers. Six benchmark databases are used for training and testing of the proposed algorithm and comparison is made with 12 existing schemes. The features are carefully crafted and are extracted in different scales and color spaces as opposed to existing approaches. A stacked ensemble of Gaussian process regression is trained and validated in terms of R-squared, RMSE, PLCC, SROCC, and KROCC. The cross-dataset evaluation of the model is done using two experiments. Comparison of the proposed approach with existing approaches has been performed based on RMSE, PLCC and SROCC and values of 0.0552, 0.9802 and 0.9776 are achieved respectively for CSIQ database. The proposed model has demonstrated superior performance to the state-of-the-art for six databases and comparable performance with *Live in the Wild* challenge database.

*5.1. Future Recommendation*

The future work will focus on improvement in the proposed algorithm by introducing features focusing on quality aware features. Synthetic scoring can be used to train deep learning models to cope the limitation of subjectively scored image databases. Features extracted from these deep models can be combined with NSS features to provide a hybrid feature set for model building.

The proposed algorithm (PIQI) can be used to evaluate the performance of several image processing algorithms such as image compression [50], image dehazing [51, 52], image denoising [53], medical image enhancement [54] and duplicate removal [55] based on perceptual quality.